\documentclass[12pt]{article}

\usepackage{scicite}
\usepackage{times}
\usepackage{amsmath,amssymb,amsopn,amstext,amsfonts, amsthm, amssymb}
\usepackage{graphicx}
\usepackage{caption}
\usepackage{booktabs}
\usepackage{multirow}
\usepackage{makecell}

\graphicspath{{../figures/}}
\DeclareGraphicsExtensions{.png,.jpg}

\newenvironment{sciabstract}{%
	\begin{quote} \bf}
	{\end{quote}}

\title{Technical Report:\\
	Learning to Plan Maneuverable and Agile Flight Trajectory with Optimization Embedded Networks}

\author
{Zhichao Han$^{\ast,1,2}$, Long Xu$^{\ast,1,2}$, Liuao Pei$^{1,2}$ and Fei Gao$^{1,2}$\\
	\\
        \normalsize{$\ast$ Equal contribution}\\ 
	\normalsize{$^{1}$Institute of Cyber-Systems and Control,}\\
	\normalsize{College of Control Science and Engineering,}\\
        \normalsize{Zhejiang University, Hangzhou 310027, China.}\\
	\normalsize{$^{2}$Huzhou Institute of Zhejiang University, Huzhou 313000, China.}\\
	\normalsize{Corresponding author: Fei Gao Email: fgaoaa@zju.edu.cn
 }
}

\date{}

\begin{document}


	\baselineskip24pt


	\maketitle

\section*{Abstract}
\begin{sciabstract}
In recent times, an increasing number of researchers have been devoted to utilizing deep neural networks for end-to-end flight navigation. This approach has gained traction due to its ability to bridge the gap between perception and planning that exists in traditional methods, thereby eliminating delays between modules. However, the practice of replacing original modules with neural networks in a black-box manner diminishes the overall system's robustness and stability. It lacks principled explanations and often fails to consistently generate high-quality motion trajectories. Furthermore, such methods often struggle to rigorously account for the robot's kinematic constraints, resulting in the generation of trajectories that cannot be executed satisfactorily.
In this work, we combine the advantages of traditional methods and neural networks by proposing an optimization-embedded neural network. This network can learn high-quality trajectories directly from visual inputs without the need of mapping, while ensuring dynamic feasibility. Here, the deep neural network is employed to directly extract environment safety regions from depth images. Subsequently, we employ a model-based approach to represent these regions as safety constraints in trajectory optimization. Leveraging the availability of highly efficient optimization algorithms, our method robustly converges to feasible and optimal solutions that satisfy various user-defined constraints.
Moreover, we differentiate the optimization process, allowing it to be  trained as a layer within the neural network. This approach facilitates the direct interaction between perception and planning, enabling the network to focus more on the spatial regions where optimal solutions exist. As a result, it further enhances the quality and stability of the generated trajectories.
\end{sciabstract}

\section*{Introduction}
\label{sec:Introduction}

Unmanned aerial vehicles (UAVs) have gained widespread adoption in various societal domains, such as aerial photography, exploration, and search and rescue, due to their compact hardware design and agile maneuverability. Efficient and robust navigation modules play a crucial role in achieving UAV autonomy, attracting significant attention from both academia and industry.

Traditional navigation tasks can be divided into perception and motion planning. Perception modules process raw data from sensors such as depth cameras to construct an occupancy map and derive environment representations conducive to motion planning, such as Euclidean signed distance fields (ESDF)~\cite{han2019fiesta} or neural radiance fields (NeRF)~\cite{adamkiewicz2022vision}. Motion planning modules, on the other hand, utilize the constructed map along with robot state, kinematics, and obstacle avoidance constraints to compute energy-minimizing trajectories with time regularization. While traditional navigation strategies offer intuitive engineering solutions and theoretical completeness and interpretability, they suffer from delays in the modular decomposition framework, adversely affecting agile flight capabilities of quadcopters~\cite{loquercio2021learning}. Moreover, the layered framework can cause a lack of interconnections between sub-modules, making the flight performance susceptible to sensor noise and often requiring manual fine-tuning of numerous parameters by engineers~\cite{han2024neupan}.

In recent years, there has been increasing research interest in directly fusing perception and planning modules into a single neural network. This end-to-end pipeline enables learning motion directly from sensor data, bypassing explicit mapping. However, such strategies result in a black-box navigation system, making debugging challenging. Furthermore, attaining a favorable balance between dynamic feasibility, obstacle avoidance, and high-quality trajectory generation places a significant burden on the network. 
For instance, due to physical platform or task constraints, it is often desired to impose dynamic constraints such as maximum velocity and acceleration constraints on the trajectories.
To address these issues, researchers often resort to designing complex strategies for the network, which can impact optimality and still fail to guarantee dynamic feasibility of the trajectories.

In this work, we combine the strengths of traditional trajectory optimization and neural networks to develop an end-to-end visual navigation system capable of generating trajectories directly from depth information without the need for explicit mapping. A key feature of our approach, compared to conventional learning-based motion planning algorithms, is the incorporation of numerical optimization within the neural network, coupled with joint training. This method alleviates the burden on the network, enhances the interpretability of the system, and ensures optimal and dynamically feasible motion trajectories. Moreover, our method is scalable and allows for the inclusion of additional user-defined constraints without the need for retraining.
In our technical approach, we utilize the neural network to directly extract safe guidance regions from the depth information, which are then transformed into geometric spatial constraints considered during trajectory optimization. The trajectory optimization process incorporates these constraints as safety boundaries while simultaneously integrating user-specified dynamic constraints, resulting in efficient convergence ($\sim$1ms) to high-quality trajectories. This ensures that the quadcopter remains maneuverable and agile even in complex environments. Unlike conventional learning-based black-box navigation systems, our optimization algorithm, enabled by a clear mathematical model, robustly converges to optimal solutions within the feasible topological space generated by the network. Furthermore, by making numerical optimization differentiable, it can be modeled as a layer and trained jointly with the neural network. This allows the gradient of the evaluation loss of the trajectories to be directly backpropagated to the network, enabling the network to focus on the spatial regions where the optimal trajectories reside. Moreover, to ensure sufficient exploration of the environment during actual flight, we introduce motion primitives within the network. The network outputs the selection probabilities for each motion primitive, and based on these probabilities, safe and feasible spaces are assigned to certain primitives.  In practical applications, we can parallelly perform trajectory optimization within the safe spaces represented by high-probability motion primitives, and select the optimal trajectory as the execution plan.
The main contributions of this paper can be summarized as follows:
\begin{itemize}
	\item We have designed a lightweight neural network capable of directly identifying valuable motion primitives from depth information, which are then used to generate the necessary safety spatial constraints for subsequent numerical optimization.
	\item By making trajectory optimization differentiable, we treat it as a layer and train it jointly with the network. This seamless integration enables the neural network to evolve directly towards optimal trajectories, eliminating any gaps between the optimization and learning processes.
	\item 
	By leveraging the strengths of neural networks and numerical optimization, we propose a high-quality, map-free planning approach. This approach enables the instantaneous generation of optimal and safe trajectories, strictly adhering to dynamic constraints.

\end{itemize}
\section*{Related Work}
\label{sec:Related Work}
\subsection*{Classical Motion Planning Algorithms}
Gradient-based motion planning~\cite{zhou2021raptor,zhou2019robust, zhou2020ego,zhou2020robust, gao2020teach,tordesillas2021mader } is widely adopted for generating local trajectories for UAVs, treating the problem as constrained nonlinear optimization.
Such methods typically require the explicit construction of the environment through depth information, followed by the manual design of strategies to extract safety constraints for trajectory optimization.
Euclidean Signed Distance Fields (ESDF) are widely used for modeling safety constraints, as they provide signed distance and gradient information from any grid point to obstacles within the space~\cite{zhou2021raptor, zhou2019robust}. However, constructing ESDF incurs additional computational costs and involves a trade-off between efficiency and accuracy, as higher resolutions exponentially increase computation and memory requirements.
Zhou et al.~\cite{zhou2020ego} proposed the well-known ego planner, which avoids ESDF construction by continuously generating safe guidance paths within an iterative framework to provide obstacle avoidance gradients. However, this method lacks convergence guarantees and may be prone to getting trapped in unsafe local minima, especially in complex environments. Furthermore, using guidance paths to deform trajectories deviates from the original trajectory optimization problem formulation, affecting optimality.
Corridor-based methods~\cite{gao2020teach,tordesillas2021mader} have also gained popularity in the field of local motion planning. These methods extract feasible convex hulls from the environment point cloud using geometric computations to model safety constraints as linear or cone constraints. However, these methods require an additional collision-free path to provide seed points for the convex hull. Such paths are often obtained using low-dimensional search algorithms like A* or hybrid A*. These search algorithms often do not consider the robot's higher-order kinematics, resulting in convex hulls that are not conducive to generating maneuverable and dynamically feasible trajectories.
\subsection*{Learning-Based Motion Planning Algorithms}
Learning-based methods~\cite{loquercio2021learning,allen2019real,chou2021uncertainty, yang2023iplanner,roth2023viplanner,kulkarni2024reinforcement,jacquet2024n,han2024neupan} have emerged as promising approaches in the field of local planning, eliminating the need for explicit mapping and reducing latency. Loquercio et al.~\cite{loquercio2021learning} leveraged deep convolutional neural networks to learn flight trajectories from depth images, using human pilot trajectories as supervision. However, this method requires high-quality and large-scale datasets.  Kulkarni et al.~\cite{kulkarni2024reinforcement} employed reinforcement learning for end-to-end navigation and enhanced safety through a custom depth collision encoder. These methods heavily rely on the capabilities of neural networks and lack principled guarantees regarding kinematic feasibility and trajectory optimality.
Recently, some approaches have combined networks with numerical optimization. For instance, a particular work~\cite{jacquet2024n} learned collision probabilities for any point in space using a network, which were further modeled as safety constraints in trajectory optimization. Similarly, another work~\cite{han2024neupan} addressed finer obstacle avoidance by modeling the robot's shape as a convex hull and predicting the signed distance between the hull and the nearest obstacle using a neural network. Although these works integrate networks and optimization for robot navigation, a significant difference compared to our approach is that the network and optimization are independent components in the aforementioned works. In contrast, our algorithm incorporates differentiable optimization as part of the network training process. Consequently, our method facilitates the evolution of the network towards directions beneficial for subsequent optimization, ultimately generating higher-quality solutions while ensuring maneuverability and agile flight.

Recent works have also applied the concept of bilayer optimization. Chen et al.~\cite{chen2024ia} propose the IA* algorithm, which differentiates the traditional A* search algorithm to embed it within a neural network for training. However, the paths generated by this algorithm consist of discrete grid points, making it unsuitable for direct tracking by high-speed drones. Additionally, this algorithm requires a global map as input, unlike our approach which directly utilizes visual information. Similar to our work, Iplanner~\cite{yang2023iplanner} is an end-to-end local planner that outputs trajectories directly from depth images and has been widely applied to quadruped robots. In its implementation, this approach uses closed-form cubic splines to interpolate the points output by the network, approximating a suboptimal solution to the original trajectory planning problem considering obstacle avoidance. While this approach is lightweight, it does not guarantee strict adherence to user-defined constraints, thereby limiting its application in scenarios where constraints must be strictly met. For example, for quadrotors, we need to impose thrust constraints, which are crucial for ensuring high-speed and stable flight.

\section*{Methodology}
\label{sec:method}
\subsection*{End-to-End Navigation System Overview}
In this work, we employ a unique class of trajectories called $MINCO$~\cite{wang2022geometrically} to represent flight trajectories $\xi$.  $MINCO$ is a special multi-piece polynomial representation  parameterized by piece durations $\mathbf{T} = [T_1, T_2, ..., T_N]^T\in\mathbb{N+}$  and waypoints $\mathbf{q} = [q_1, q_2, ..., q_{N-1}]\in\mathbb{R}^{3\times (N-1)}$, where $N$ is denoted as the number of trajectory pieces.
This compact representation naturally satisfies the state constraints at the start and end points, as well as the high-order continuity of adjacent polynomials at the waypoints, without the need for additional constraints.
Building upon this representation, the trajectory optimization is formulated as the minimization of control energy with first-order temporal regularization, and can be expressed as follows:
\begin{align}
&\min_{\mathbf{q}, \mathbf{T}} J = \int_{0}^{||\mathbf{T}||_1}(\xi^{(u)}(t))^{\rm T}\mathrm{W}\xi^{(u)}(t)dt+\rho ||\mathbf{T}||_1 \label{eq:objective function} \\
&s.t. ~~~\mathcal{G}(\xi(t), \xi^{(1)}(t),...,\xi^{(u)}(t))\leq 0, \forall t  \in [0, ||\mathbf{T}||_1] \label{eq:dynamic} \\
&~~~~~~~~\xi(t) \in \mathcal{F}, \forall t  \in [0, ||\mathbf{T}||_1],\label{eq:flight}
\end{align}
where $\mathrm{W}$ is a positive definite energy weight matrix. $\rho$ is the temporal regularization weight and $u$ represents the dimension of the control variable.
$\mathcal{G}$  represents pre-defined kinematic constraints, which are specifically formulated based on user's requirements and the robot's dynamics. Additionally, Eq. (\ref{eq:flight}) represents the obstacle avoidance constraints based on the flight corridor $\mathcal{F}$.
Typically, this constraint can be accurately modeled as a linear or conic constraint with respect to the robot's position coordinates.
Due to advancements in the field of optimization, once the trajectory optimization Eq. (\ref{eq:objective function}-\ref{eq:flight}) is fully formulated and modeled, mature gradient-based numerical methods are available that can efficiently converge to high-quality solutions at low computational cost. However, accurately extracting feasible spaces from complex and cluttered environments during actual flight poses a significant challenge. As mentioned in Sect. I, it involves multiple modules such as sensor data processing, mapping, graph search, etc., and is susceptible to noise, making this aspect the Achilles' heel of the entire navigation system. Therefore, in this work, we intuitively explore the utilization of neural networks to learn flight corridors directly from depth images, without the need for explicit mapping. Essentially, we employ neural networks to learn safety constraints within the trajectory optimization process. 

\begin{figure}[t]
    \centering
    \includegraphics[width=1.0\columnwidth]{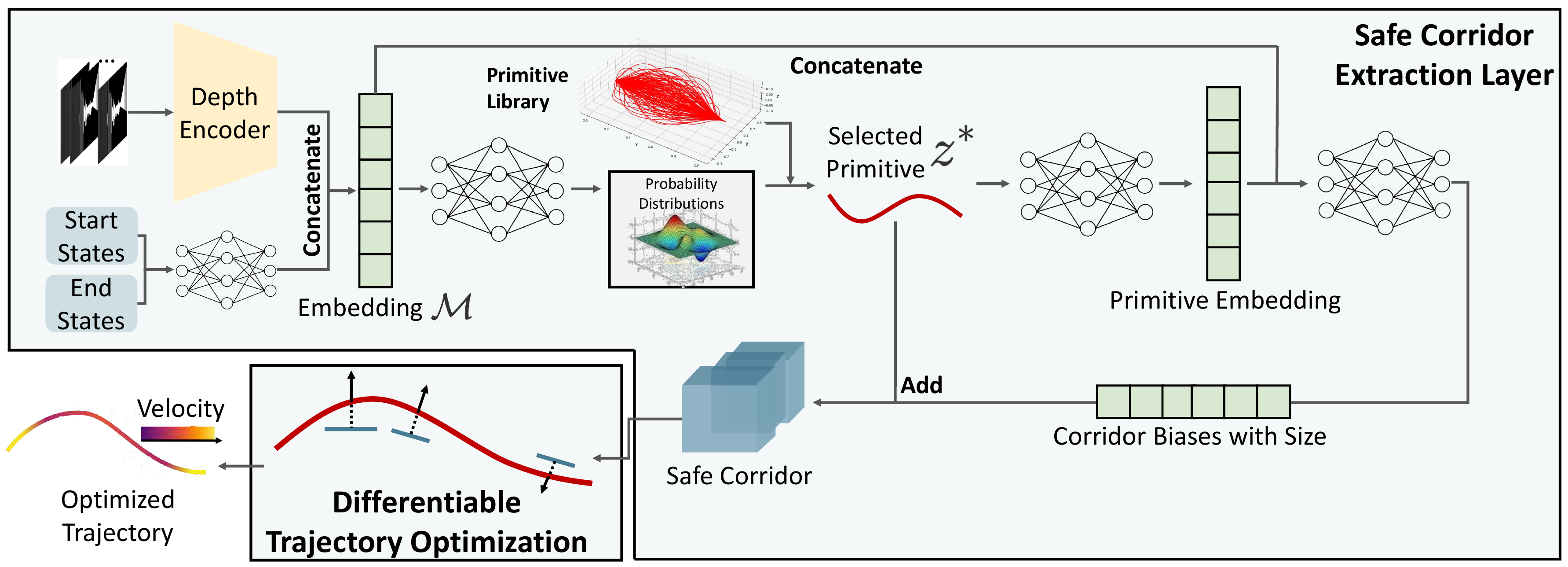}
	\caption{Navigation framework.}\label{fig:nn}
\end{figure}

Our navigation framework, as illustrated in Fig.~\ref{fig:nn}, can be broadly divided into a network layer for extracting flight corridors and a differentiable trajectory optimization layer.
The network layer takes inputs such as depth measurements, the current state of the robot, and the target point's position to output flight corridors.
Subsequently, the trajectory optimization layer plans a high-quality spatial-temporal optimal trajectory constrained within the flight corridors, while strictly adhering to specified dynamic constraints.
Furthermore, the inputs and outputs of each module, such as the target point, flight corridors, and planned trajectory, are normalized with respect to the robot's current body frame. This normalization enhances the generalization capability of the model and reduces the dependence of the entire navigation system on global localization.
In the following sections, we will first introduce the structural design of the network layer and provide a detailed description of its output. 
Then, we discuss the gradient propagation of the optimal solution generated by the trajectory optimization layer with respect to spatial constraints, which forms the foundational basis for jointly training the embedded optimization within the network.
\subsection*{Learning-Based Safe Corridor Extraction Layer}
Before delving into the specific details of the network structure, we first instantiate its output representation. For continuous-time constraints Eq. (\ref{eq:flight}),  similar to the approach \cite{han2023efficient}, we discretize each piece of the polynomial into $\lambda$ constraint points. Subsequently, we control the entire trajectory by imposing constraints at these constraint points:
\begin{align}
\xi(t) &\in \mathcal{F}, \forall t  \in [0, ||\mathbf{T}||_1] \iff  \nonumber \\
\xi_i(\frac{j}{\lambda T_i}) &\in \mathcal{F}_{i,j}^{\phi} \forall i \in [1,...,N],\forall j \in [1,...,\lambda],\label{eq:discrete_f}
\end{align}
where $\phi$ is the parameters of the neural network.
The physical significance of Eq. (\ref{eq:discrete_f}) lies in the fact that the neural network needs to assign a safety convex hull to each constraint point along the trajectory. Consequently, the network is required to output a total of $N\lambda$ convex hulls. Moreover, we would like to emphasize that the network has the theoretical capability to output convex hulls of arbitrary shapes. However, for the sake of simplicity and ease of understanding, we define each convex hull as a cube parameterized by its center and length.

Here, we design a novel neural network based on motion primitives, which enables us to fulfill the aforementioned requirements.
This network initially employs deep convolutional and fully connected layers to extract features from depth images and the initial and final states of the robot. These features are fused to obtain a latent representation, denoted as $\mathcal{M} $.
Subsequently, $\mathcal{M}$ is further processed through a neural network to output a probability distribution over a pre-built library of motion primitives $\zeta$. It is worth mentioning that each motion primitive $z$, in order to align with the subsequent corridor parameters, is represented by $N\lambda$ points.
Furthermore, the selected motion primitive $z^\ast$ and $\mathcal{M}$ are jointly fed into the final corridor generation module. One of its roles is to refine the motion primitives for improved accuracy, with the modified point coordinates serving as the centers of the safety cubes. Additionally, this module is responsible for assigning the corresponding length to each cube.
It is worth noting that the advantages of this motion-primitive-based structure are evident in at least two aspects. Firstly, compared to directly regressing the final cube centers, our approach first obtains a probability distribution and selects better motion primitives. This process can be modeled as a classification problem, which reduces the network's burden and facilitates learning and convergence. Secondly, during practical deployment, we have the flexibility to select multiple motion primitives based on their probabilities. This allows for parallel optimization of multiple trajectories, thereby enhancing the robot's exploration capabilities in diverse environmental topologies. Simultaneously, it also increases the system's fault tolerance.

Regarding the construction of the motion primitive library, in our practical experiments, we recorded tens of thousands of trajectories using classical navigation algorithms. These trajectories are uniformly discretized into $N\lambda$ points and transformed into the local body coordinate system. To eliminate unnecessary duplicate motion primitives and limit the size of the library, we normalize the endpoint distances and directions for all motion primitive data. Finally, we employ the k-means algorithm to cluster the processed dataset and collect approximately 100 elite motion primitives as the library.
\subsection*{Differentiable Numerical Optimization Layer}
In this section, we  discuss how to make the optimization process differentiable, allowing us to backpropagate the gradients of the loss applied to the trajectory onto the network parameters during training.
For simplicity, we denote the optimization variables as $x = (\mathbf{q}, \mathbf{T})$. Then, the nonlinear optimization problem Eq. (\ref{eq:objective function}-\ref{eq:flight}) with inequality constraints can be generally reformulated as follows:
\begin{align}
&\min_{x} J = J(x) \label{eq:obj2} \\
&s.t. ~~~F(x, \phi)\leq 0. \label{eq:cons2} 
\end{align}
Here, $F$ represents a general formulation that encompasses the original constraints Eq. (\ref{eq:dynamic},\ref{eq:flight}), and $\phi$ denotes the neural network parameters.
Assuming $x^*$ is the optimal solution to this optimization problem, and $\mathcal{L}$ is the evaluation loss applied to the trajectory during training, the gradient of the neural network can be computed as follows:
\begin{align}
\nabla\phi \mathcal{L} =\nabla_\phi x\nabla_x \mathcal{L}  \label{eq:propogate}
\end{align}
Generally, the term $\nabla_x \mathcal{L}$ can be analytically computed. Therefore, our focus now shifts to discussing the estimation of $\nabla_\phi x$.
Due to the usage of gradient-based numerical solvers, a intuitive method for estimating parameter gradients, known as unrolling~\cite{finn2017model,bhardwaj2020differentiable,pearlmutter2008reverse,zhang2010multi,han2017alternating}, involves maintaining the entire computational graph throughout the iteration process. However, this approach poses significant challenges in terms of memory usage and efficiency, particularly when dealing with complex problem formulations. Moreover, it may also face issues related to gradient divergence or vanishment.
In this work, we assume the efficient attainment of the optimal solution $x^*$ to the problem, and thus employ the implicit function differentiation theorem from Dontchev and Rockafellar \cite{dontchev2009implicit}. This method relies on leveraging the first-order optimality condition~\cite{amos2017optnet, agrawal2019differentiable} of the optimization problem to analytically estimate the gradients of the parameters without the need for explicit unrolling of the entire iteration process. 
The Lagrangian function for this optimization problem is as follows:
\begin{align}
L(x,\lambda) = J(x)+\lambda^TF(x,\phi).
\end{align}
Then, the corresponding KKT (Karush-Kuhn-Tucker) conditions are as follows:
\begin{align}
    \nabla_x J + \nabla_xF \lambda^* &= 0,\nonumber\\
    D(\lambda^*)F(x^*,\phi) &= 0, \nonumber\\
    F(x^*,\phi) &\le0,\nonumber\\
    \lambda^*&\ge0,
    \label{eq:kkt}
\end{align}
where $D(\cdot)$ denotes a diagonal matrix from a vector.
Then, we apply the total differential operator $\mathrm{d}$ to the equations in KKT conditions:
\begin{align}
    (\nabla_{x,x}J
    +(\nabla_{x,x}F \lambda^*)^{\mathrm{T}})\mathrm{d}x+\nabla_xF\mathrm{d}\lambda+(\nabla_{x,\phi}F\lambda^*)^{\mathrm{T}}\mathrm{d}\phi &= 0,
    \nonumber\\
    D(F)\mathrm{d}\lambda+D(\lambda^*)\nabla_xF\mathrm{d}x
    +D(\lambda^*)\nabla_\phi F \mathrm{d}\phi &= 0.\label{eq:kkt}
\end{align}
Subsequently, Eq. (\ref{eq:kkt}) is further transformed into a compact matrix form:
\begin{align}
\begin{bmatrix}
\mathrm{d}x \\
\mathrm{d}\lambda
\end{bmatrix}
= -
\begin{bmatrix}
\nabla_{x,x}J +(\nabla_{x,x}F \lambda^*)^{\mathrm{T}})
 & \nabla_{x}F
\\
D(\lambda^*)\nabla_xF & D(F)
\end{bmatrix}^{-1}
\begin{bmatrix}
(\nabla_{x,\phi}F\lambda^*)^{\mathrm{T}}
\\
D(\lambda^*)\nabla_\phi F
\end{bmatrix}\mathrm{d}\phi.
\end{align}
By solving this system of equations, we can analytically obtain the desired Jacobian matrix $\nabla_\phi x$, which in turn allows us to derive the final parameter gradients $\nabla\phi \mathcal{L}$.

\section*{Results}
\subsection*{Ablation Experiments}
To verify the effectiveness of embedding the numerical optimization into the neural network, we conducted ablation experiments in a test set with more than ten thousand items of data, quantitatively comparing the safety of guidance regions extracted by the neural network and the energy consumption of the optimized trajectories, as shown in Table \ref{tab:Ablation}. We consider it unsafe when the edge of the guidance region touches any obstacle. Besides, the integral with respect to time of the square of the jerk of the trajectory is utilized to measure the energy consumption. 

\begin{table}[h]
	\small
	\centering
	\renewcommand\arraystretch{2.0}
        
	\caption{Ablation Experiments. \protect\label{tab:Ablation}}
	\begin{tabular}{l|cc}
		\hline
		  & Safety Ratio $\uparrow$  &  Avg. Energy $\downarrow$\\
		\hline
		w/o Opt. &82.1\%  &7.262 \\
		\hline
		  w/ Opt. &85.2\%  &5.285 \\
		\hline
	\end{tabular}
\end{table}

Since we remove the imitation of the ground truth and add the loss function minimizing the energy of the trajectory after embedding numerical optimization in the neural network, the safe regions extracted by the network are quickly tuned towards the direction of reducing energy consumption. At the same time, as mentioned in the previous section, with backpropagation of the energy loss function, we can remove the requirement for the safe region to be as large as possible, which makes the security constraints of the region easier to satisfy, allowing the network to focus more on containing the optimal solution, thus resulting in a higher safety ratio.

\begin{figure}[t]
    \centering
    \includegraphics[width=1.0\columnwidth]{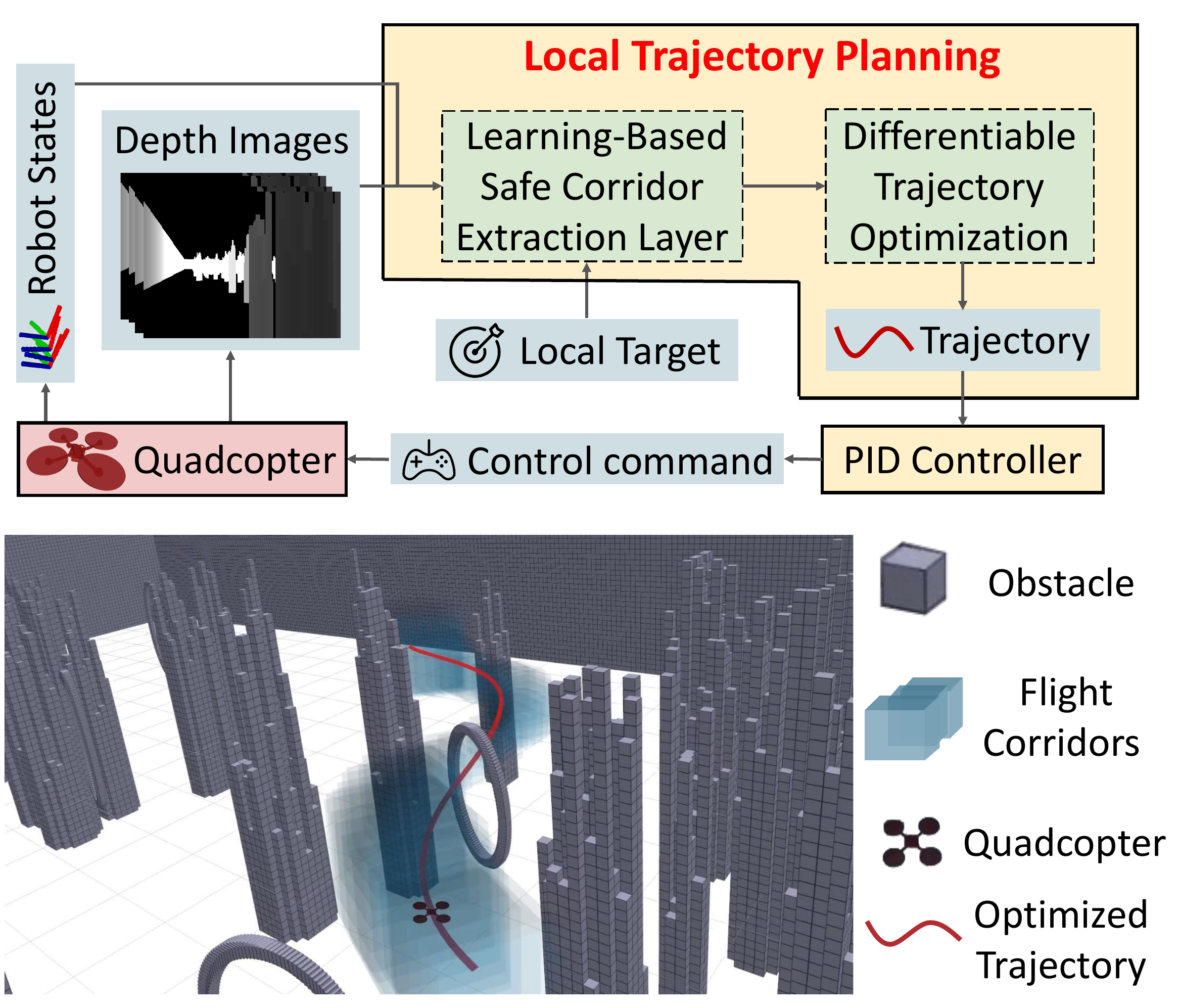}
	\caption{Simulation experiments and system framework.}\label{fig:sim-exp}
\end{figure}

\subsection*{Simulation Experiments}
We deploy the algorithm on a simulated quadcopter, conducting experiments in an environment with dynamics simulation, as shown in Fig.~\ref{fig:sim-exp}. The size of the simulation environment is 50m $\times$ 50m, which contains randomly generated obstacles in the form of columns and loops. With known localization, we require the quadcopter to start from a random location at the edge of the map and continuously replan to traverse the entire unknown environment.

In addition, we quantitatively compare the proposed method with the state-of-the-art algorithm Ego-planner\cite{zhou2020ego} in terms of objective function value and total processing latency, with more than four thousand comparative tests. All simulations are run on a desktop with an Intel i9-10900K CPU and a Nvidia GeForce RTX3070Ti GPU.

\begin{table}[h]
	\small
	\centering
	\renewcommand\arraystretch{1.5}
	\caption{Algorithm Comparisons. \protect\label{tab:sim}}
	\begin{tabular}{l|cc}
		\hline
            & \makecell[l]{Objective Fu\\-nction Value} $\downarrow$ & \makecell[l]{Total Proces-\\sing Latency} $\downarrow$\\
            \hline
		proposed &59.90 $\pm$ 8.71 &3.49 $\pm$ 0.73ms\\
		\hline
		  Ego-planner\cite{zhou2020ego} &61.65 $\pm$ 11.73 &24.33 $\pm$ 16.58ms\\
		\hline
	\end{tabular}
\end{table}

In Table \ref{tab:sim}, our approach achieves better performance than Ego-planner\cite{zhou2020ego}. The use of collision-free trajectories as ground truth to supervise and training based on backpropagation of the results of trajectory optimization make the neural network give safety regions that more easily contains the optimal solution. Thus, the proposed algorithm has a greater advantage in terms of objective function value. Compared to classical pipelines, neural networks have more stable inference times and outputs, bringing low standard deviations. Also, the end-to-end pipeline and parallel GPU-based reasoning make it have lower latency. 

\section*{Discussion}
In this work, we propose a novel planning framework for maneuverable and agile flight of quadrotors. Using an embedded optimized neural network, we plan trajectories directly from visual measurements without the need for explicit mapping, while being able to guarantee the dynamical feasibility of the trajectories. Benefiting from the differentiable trajectory optimization, the burden on the neural network is relieved to more easily extraction of safe regions containing optimal solutions. Simulation experiments prove the efficiency of the pipeline compared to state-of-the-art method. In the future, more quantitative experiments will be carried out in high speed flights in complex environments, where the success rate of classical methods will plummet due to high latency. We will also further optimize the proposed pipeline and conduct more comparisons with other learning-based methods. Besides, this framework will be deployed to real quadrotors for more tests.

\bibliography{scibib}
\bibliographystyle{Science}

\end{document}